\documentclass[10pt,twocolumn,letterpaper]{article}

\usepackage[pagenumbers]{cvpr}
\usepackage{xcolor}
\usepackage{booktabs,graphicx,amssymb,makecell}
\usepackage{multirow,array,colortbl}
\usepackage{tabularx}

\usepackage{tcolorbox}
\tcbuselibrary{breakable,skins}

\definecolor{cvprblue}{rgb}{0.21,0.49,0.74}
\usepackage[pagebackref,breaklinks,colorlinks,allcolors=cvprblue]{hyperref}

\def\confName{CVPR}
\def\confYear{2027}

\title{BasketTactics: Benchmarking Possession-Level Tactical Reasoning in Basketball}

\author{%
  Ya Jing\textsuperscript{1,*},
  Yangdong Gao\textsuperscript{1,*},
  Yuyang Wu\textsuperscript{1,*},
  Yuancheng Sun\textsuperscript{2},
  Shuochen Zhang\textsuperscript{3,4}\\
  Shikun Feng\textsuperscript{4,7},
  Lei Zheng\textsuperscript{5},
  Guole Zhu\textsuperscript{5},
  Jiacheng Yan\textsuperscript{6},
  Delong Zhao\textsuperscript{1}\\
  Yunhao Zhao\textsuperscript{1},
  Yiheng Zhu\textsuperscript{4,7,\ensuremath{\dagger}},
  Lifang Wu\textsuperscript{1,\ensuremath{\dagger}},
  Ming He\textsuperscript{1}\\[0.5em]
  \textsuperscript{1}Beijing University of Technology\\
  \textsuperscript{2}Beijing Academy of Artificial Intelligence\\
  \textsuperscript{3}Tianjin University\quad
  \textsuperscript{4}Zhongguancun Academy\\
  \textsuperscript{5}Beijing Sport University\quad
  \textsuperscript{6}North China University of Technology\\
  \textsuperscript{7}Zhongguancun Institute of Artificial Intelligence%
}

\begin{document}
\maketitle
\begingroup
\makeatletter
\let\@footnotetext\H@@footnotetext
\makeatother
\renewcommand{\thefootnote}{}%
\footnotetext{\raggedright
  \textsuperscript{*}Equal contribution.\quad
  \textsuperscript{\ensuremath{\dagger}}Corresponding author.\par
  Ya Jing (\href{mailto:jingya004@126.com}{\nolinkurl{jingya004@126.com}}),
  Yangdong Gao (\href{mailto:gyd2389@163.com}{\nolinkurl{gyd2389@163.com}}),
  Yuyang Wu (\href{mailto:yuyang_wu2003@163.com}{\nolinkurl{yuyang_wu2003@163.com}}),
  Yiheng Zhu (\href{mailto:zhuyiheng@zgci.ac.cn}{\nolinkurl{zhuyiheng@zgci.ac.cn}}),
  Lifang Wu (\href{mailto:lfwu@bjut.edu.cn}{\nolinkurl{lfwu@bjut.edu.cn}}).}
\endgroup
\begin{abstract}
Understanding basketball tactics requires reasoning about how coordinated actions and defensive responses shape a possession.
Such reasoning also requires assessing tactical execution separately from scoring outcomes, since a favorable opportunity may end in a miss while a disrupted play may still score.
However, existing sports benchmarks provide limited assessment of these requirements and rarely pair anticipation with retrospective analysis of the same possession.
We introduce BasketTactics, a benchmark comprising 1,968 basketball possessions and 39,954 question-answer pairs.
Given video prefixes ending at task-specific observation cutoffs, models predict actions, tactics, and possession outcomes, identify participating players, and justify predictions using only observed evidence.
Using complete videos of the same possessions, models reconstruct offensive and defensive execution, assess tactical objective attainment separately from scoring outcomes, and provide diagnoses and adjustment recommendations.
Evaluation of ten zero-shot multimodal large language models and one fine-tuned variant reveals difficulties in anticipating possession development and reconstructing offensive and defensive execution.
Error analysis identifies incomplete evidence in predictive explanations and visual perception errors in retrospective reviews.
BasketTactics advances evaluation of tactical reasoning by connecting anticipation with execution analysis beyond the final score.
\end{abstract}    
\section{Introduction}
\label{sec:intro}

Recent advances in multimodal large language models (MLLMs) have enabled models to describe and explain video content in response to natural-language queries~\cite{qwen3vl,videollama3,internvl3}.
In team sports, an important question is whether these models can explain how interactions among players shape the development of a play.
Such understanding could support coaching and game review by helping interpret player coordination, assess defensive responses, and identify opportunities for tactical adjustment.
Basketball provides a concrete setting for studying tactical reasoning because the significance of a pass, screen, or off-ball movement depends on the positions, movements, and responses of multiple players.
These interactions unfold over a possession, as an initial setup develops through successive actions and adjustments~\cite{cervone2016multiresolution}.
A complete offensive possession is therefore a natural unit for evaluating tactical reasoning, as it captures how individual actions and defensive responses create or limit scoring opportunities.

Existing benchmarks already assess several capabilities relevant to this goal, including temporal reasoning, causal reasoning~\cite{nextqa}, and the visual grounding of answers~\cite{nextgqa}.
Sports-specific evaluations further examine professional action understanding~\cite{sportsqa}, rule application~\cite{sportu}, and tactical explanations~\cite{sportr}.
Comprehensive sports benchmarks also evaluate capability integration~\cite{soccerbench}, with BasketballBench~\cite{basketballbench} examining how event understanding is linked to player identification in possession-level clips.
For tactical analysis, TRACE~\cite{tennisvar} links tactical answers to temporally ordered sequences of supporting strokes in tennis, and SportD~\cite{sportd} evaluates strategic action selection in soccer using estimated action values.
However, evaluations that pair anticipation with retrospective analysis of the same basketball possession and assess tactical objective attainment separately from scoring outcomes remain limited.

Evaluating possession-level tactical reasoning requires clear observation boundaries and explicit criteria for judging execution.
For prediction, task-specific observation cutoffs must occur before the queried events or results become observable.
The subsequent continuation determines the reference answers, but supporting explanations must rely exclusively on the observed video prefix, including visible positions, movements, and interactions.
In retrospective analysis, the full possession is available, but the final score alone does not establish whether tactical objectives were achieved.
A screen may create an open shot that misses, whereas a defensive rotation may force a contested attempt that nevertheless scores.
Tactical review must therefore examine how the preceding sequence of offensive and defensive interactions created or restricted opportunities, assessing objective attainment separately from the final score.

\begin{figure*}[t]
    \centering
    \includegraphics[width=0.985\textwidth]{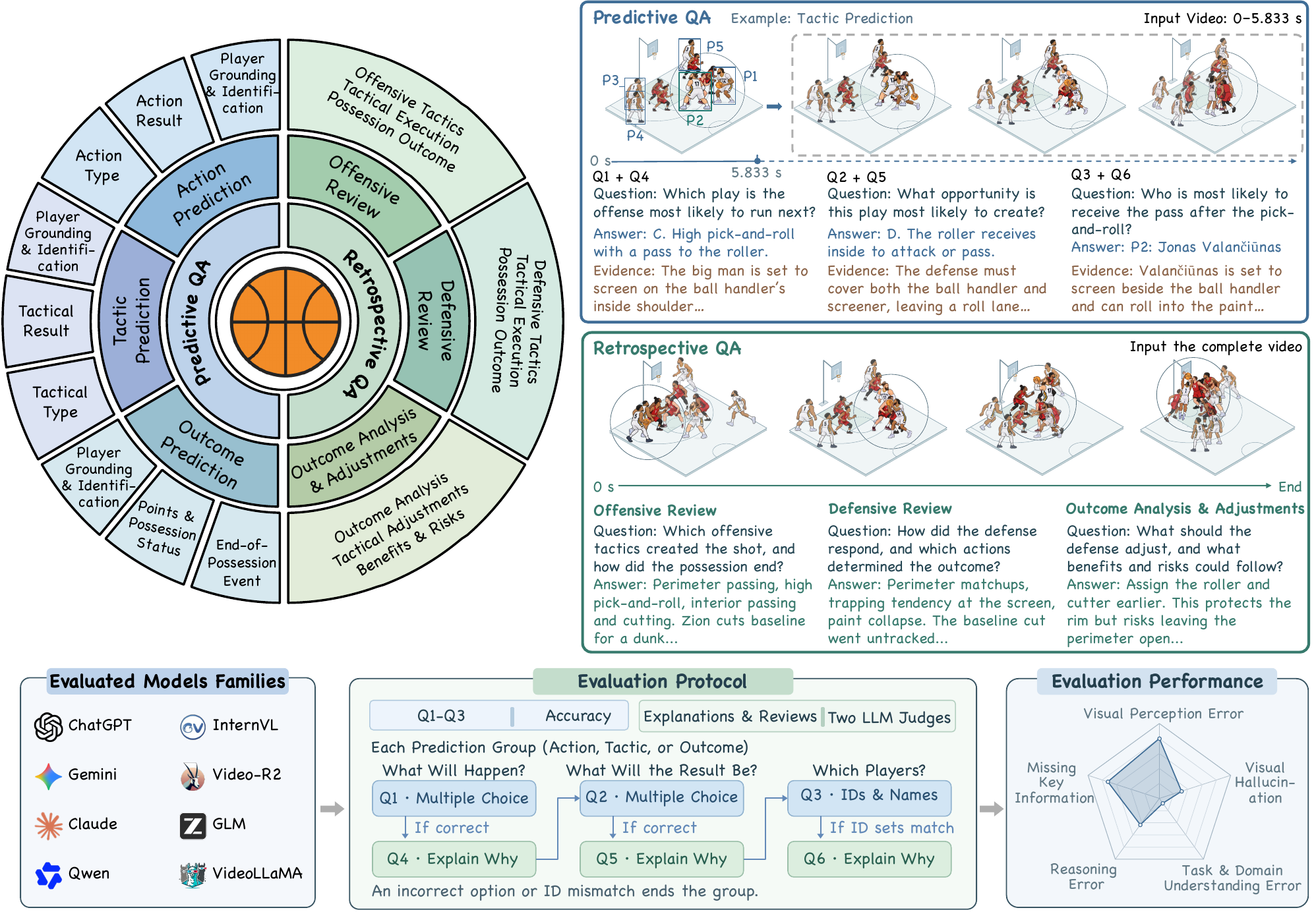}
    \caption{
    Overview of BasketTactics, including its task taxonomy, representative QA pairs, and evaluation protocol. The benchmark pairs action, tactic, and outcome prediction from video prefixes with full-possession retrospective analysis, assessing tactical objective attainment separately from realized scoring outcomes.
    }
    \label{fig:overview}
\end{figure*}

We introduce BasketTactics, a benchmark that pairs predictive and retrospective tasks on the same basketball possessions, as illustrated in Figure~\ref{fig:overview}.
Built on possession-level videos from PL-NBA~\cite{pl-nba}, it comprises 1,968 possessions and 39,954 question-answer pairs organized into six task groups covering 21 question types.
Action, tactic, and outcome prediction evaluate how models anticipate possession development from video prefixes ending at task-specific observation cutoffs.
Each group connects the predicted event and its result to participating players and supporting explanations, extending evaluation beyond the correctness of the anticipated event alone.
Bounding boxes and frame-specific IDs for visible offensive players provide explicit visual references for these event-participant associations.
The same possessions also support offensive review, defensive review, and outcome analysis using complete videos.
These tasks examine how models reconstruct tactical execution through player movements, coordination, and defensive adjustments, with tactical objective attainment and scoring opportunity quality assessed separately from the scoring outcome.
Outcome analysis extends the review to diagnoses and recommendations, requiring models to identify observable contributing factors and explain the potential benefits and risks of proposed adjustments.
Together, these tasks support evaluation of both anticipation and retrospective understanding within a shared possession context, examining whether models can identify the relevant players, explain their interactions, and assess what those interactions achieve.

We evaluate ten zero-shot MLLMs and one fine-tuned variant on BasketTactics.
Models struggle to jointly predict events, their results, and participating players, and to reconstruct offensive and defensive execution.
Error analysis identifies incomplete supporting evidence in predictive explanations and dominant visual perception errors in sampled retrospective reviews.
To probe the role of factual information, we supplement videos with chronological action records, restricted to the observation cutoff for prediction.
These records improve initial prediction accuracy and retrospective review scores across all evaluated configurations.
These findings demonstrate the diagnostic value of BasketTactics in revealing limitations in evidence use and execution reconstruction and assessing how additional factual information affects tactical reasoning performance.

Our contributions are summarized as follows:
\begin{itemize}

    \item We introduce BasketTactics, a possession-level basketball benchmark pairing prediction and retrospective analysis of each possession to evaluate how models anticipate possession development and interpret tactical execution.
    \item We develop an evaluation framework linking predicted events to their results, participants, and supporting visual evidence, while distinguishing tactical objective attainment from scoring outcomes in retrospective reviews.
    \item We evaluate a broad range of open-source and proprietary MLLMs, with error analysis and factual-context experiments revealing limitations in evidence use and execution reconstruction and assessing how additional action information supports prediction and retrospective analysis.
    
\end{itemize}
\section{Related Work}
\label{sec:related}

\subsection{General Video Understanding Benchmarks}
Video understanding benchmarks assess complementary aspects of perception, reasoning, and prediction.
Broad evaluations include temporal skills in MVBench~\cite{mvbench} and diverse domains and durations in Video-MME~\cite{videomme}.
Targeted benchmarks examine temporal and causal reasoning in NExT-QA~\cite{nextgqa}, compositional spatiotemporal reasoning in AGQA~\cite{agqa}, and answer grounding in NExT-GQA~\cite{nextgqa}.
STAR~\cite{star} includes prediction and feasibility questions grounded in observed situations, while Ego4D~\cite{ego4d} evaluates future actions and interactions from egocentric video.
BasketTactics studies these capabilities in coordinated basketball play, where interpreting an action depends on the movements and responses of multiple players.

\subsection{Sports Understanding Benchmarks}
Sports benchmarks connect visual observations with domain knowledge about actions, rules, and game context.
Sports-QA~\cite{sportsqa} examines relationships among professional actions, while SPORTU~\cite{sportu} and SportR~\cite{sportr} extend evaluation to rule application and tactical interpretation.
Understanding extended play also requires integrating evidence across time and connecting different perceptual tasks.
SportsTime~\cite{sportstime} addresses temporal integration with step-wise evidence annotations, while SoccerBench~\cite{soccerbench} and BasketballBench~\cite{basketballbench} combine multiple tasks and modalities.
BasketballBench specifically links event recognition and localization to player identification in possession-level clips.
SPRINT~\cite{sprint} further considers future hazards and their causes.
BasketTactics organizes prediction and retrospective analysis around the same basketball possession, connecting anticipated events and their participants with the subsequent development of play.

\subsection{Tactical Analysis and Decision-Making}
Tactical analysis studies how player interactions shape opportunities and inform decisions.
Structured approaches support corner-kick prediction and positional adjustments in TacticAI~\cite{tacticai}, style imitation in TacSIm~\cite{tacsim}, and trajectory forecasting and conditional simulation in GenTac~\cite{gentac}.
Video-language evaluations link tactical answers to stroke evidence in TRACE~\cite{tennisvar}, and assess action choices using estimated strategic values in SportD~\cite{sportd}.
These directions address tactical evidence and decision quality.
In particular, SportD~\cite{sportd} distinguishes action success probability from strategic value, showing that a higher success probability does not necessarily imply greater value.
BasketTactics emphasizes the distinction between tactical objective attainment and realized scoring, examining offensive-defensive execution across complete possessions to support diagnosis and adjustment recommendations.
\section{BasketTactics Benchmark}
\label{sec:benchmark}

BasketTactics comprises 21 question types across six task groups, as summarized in Figure~\ref{fig:overview}.
Action, tactic, and outcome prediction each comprise six question types, while offensive review, defensive review, and outcome analysis and adjustments each comprise one open-ended question type.

\subsection{Task Formulation}
\label{sec:task_formulation}

BasketTactics evaluates prediction and retrospective analysis at the possession level.
Predictive tasks use a video prefix ending at a task-specific observation cutoff, together with the cutoff frame annotated with player bounding boxes and frame-specific IDs.
Retrospective tasks use the full episode to analyze offensive and defensive execution and its relationship to the observed outcome.

\noindent\textbf{Predictive Tasks.}
Each prediction group pairs three objective questions about a future event, its result, and participating players with three corresponding explanation questions.
Action prediction targets the next action and its result, while tactic prediction targets the upcoming offensive tactic and its tactical result.
Outcome prediction targets the finishing action or terminal event and the resulting points and possession status.
The event and result questions (Q1-Q2) are multiple-choice, while the participant question (Q3) requires the supplied frame-specific player IDs, with player names requested as auxiliary outputs.
Objective reference answers are determined by the actual continuation.
The explanation questions (Q4-Q6) ask models to justify the corresponding objective answers using only evidence available at or before the observation cutoff.

\begin{figure*}[t]
    \centering
    \includegraphics[width=\textwidth]{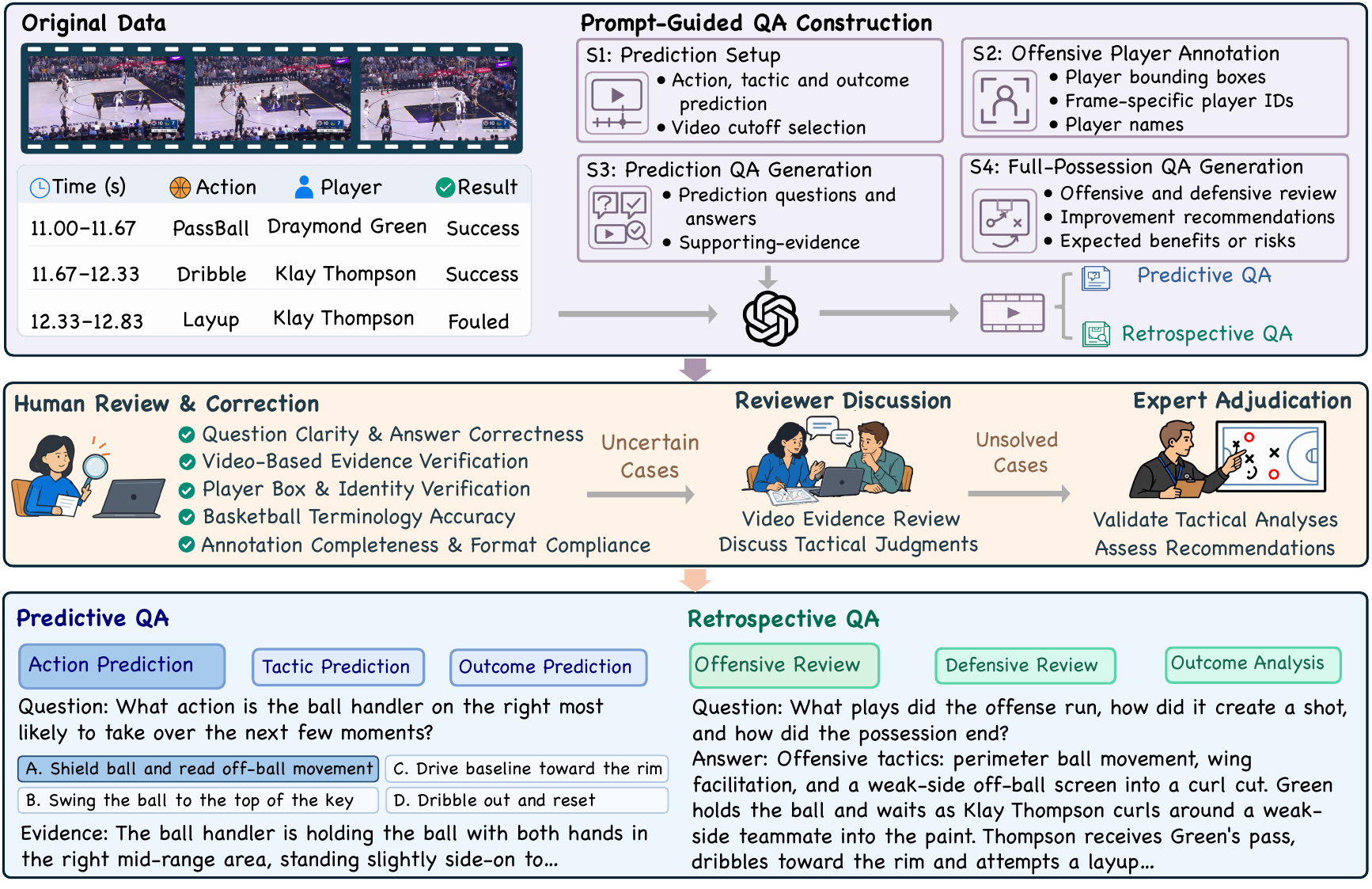}
    \caption{
    Data construction pipeline of BasketTactics. Model-assisted generation of predictive and retrospective QA pairs is followed by human review and correction, with unresolved cases referred to experts for adjudication.
    }
    \label{fig:construction_pipeline}
\end{figure*}

\noindent\textbf{Retrospective Tasks.}
Three open-ended questions examine offensive execution, defensive execution, and their relationship to the observed outcome.
Offensive review identifies the offensive tactics and reconstructs their execution through player movements, coordination, and adjustments.
Defensive review identifies the defensive strategies and explains how positioning, rotations, and responses constrain the offense.
Both reviews characterize the temporal progression and spatial organization of play.
The outcome analysis and adjustments task identifies observable factors contributing to the final outcome and asks what the side under review should retain or improve, including the potential benefits and risks of proposed adjustments.
Together, these tasks distinguish tactical objective attainment and scoring opportunity quality from the final scoring outcome.
This distinction accommodates both possessions that score despite unsuccessful tactical execution and those that create favorable opportunities but fail to score.

\subsection{Benchmark Construction}

BasketTactics combines model-assisted annotation, question filtering, and human review, as illustrated in Figure~\ref{fig:construction_pipeline}.

\noindent\textbf{Source Data and Selection.}
We build on PL-NBA~\cite{pl-nba}, which provides possession-level videos and event annotations covering action categories, temporal boundaries, player names, and outcomes.
We select clips with identifiable prediction targets and verifiable outcomes, and remove duplicates using source game and possession IDs.
Our annotation unit is a possession-based offensive episode that includes offensive rebounds and subsequent attacks within the same possession.
Episodes typically end when the offense scores, loses possession without scoring, or draws a foul.
Subsequent free throws are excluded.

\begin{table*}[t]
\centering
\caption{
Sports QA benchmarks comparison.
Player ID denotes identification by name or jersey number.
Temp., Pred., and Expl.\ denote temporal reasoning, prediction, and supporting explanations.
Tact.\ Rev.\ covers possession- or rally-level tactical analysis, with $\triangle$ indicating question-specific analysis and O+D denoting separate offensive and defensive reviews.
Exec.\ vs.\ Score distinguishes tactical execution from scoring outcomes.
Diag.\ \& Adj.\ denotes diagnosis and adjustment recommendations.
}
\label{tab:benchmark_comparison}

\begingroup
\footnotesize
\setlength{\tabcolsep}{1.25pt}
\renewcommand{\arraystretch}{1.1}

\definecolor{BTCheck}{HTML}{000000}
\definecolor{BTCross}{HTML}{000000}
\newcommand{\BTYes}{\textcolor{BTCheck}{\ensuremath{\checkmark}}}
\newcommand{\BTNo}{\textcolor{BTCross}{\ensuremath{\times}}}

\begin{tabular*}{\textwidth}{@{\extracolsep{\fill}}lcr*{9}{c}@{}}
\toprule
Benchmark
& Sport
& \#QA
& Player ID
& Temp.
& Causal
& Pred.
& Tactic ID
& Tact.\ Rev.
& Exec.\ vs.\ Score
& Diag.\ \& Adj.
& Expl. \\
\midrule

Sports-QA~\cite{sportsqa} & Multi-sport & 94,073
& \BTNo & \BTYes & \BTYes & \BTNo & \BTNo & \BTNo & \BTNo & \BTNo & \BTNo \\

SPORTU~\cite{sportu} & Multi-sport & 12,948
& \BTNo & \BTYes & \BTYes & \BTNo & \BTYes & \BTNo & \BTNo & \BTNo & \BTYes \\

SportR~\cite{sportr} & Multi-sport & 23,915
& \BTNo & \BTYes & \BTYes & \BTNo & \BTYes & \BTNo & \BTNo & \BTNo & \BTYes \\

SportsTime~\cite{sportstime} & Multi-sport & 14,326
& \BTYes & \BTYes & \BTYes & \BTNo & \BTYes & \BTNo & \BTNo & \BTYes & \BTYes \\

TRACE~\cite{tennisvar} & Tennis & 11,189
& \BTNo & \BTYes & \BTYes & \BTNo & \BTYes & $\triangle$ & \BTNo & \BTNo & \BTYes \\

SoccerBench~\cite{soccerbench} & Soccer & 9,500
& \BTYes & \BTYes & \BTYes & \BTNo & \BTNo & \BTNo & \BTNo & \BTNo & \BTNo \\

BasketballBench~\cite{basketballbench} & Basketball & 7,980
& \BTYes & \BTYes & \BTNo & \BTNo & \BTNo & \BTNo & \BTNo & \BTNo & \BTNo \\

\midrule
BasketTactics & Basketball & 39,954
& \BTYes & \BTYes & \BTYes & \BTYes & \BTYes & O+D
& \BTYes & \BTYes & \BTYes \\

\bottomrule
\end{tabular*}
\endgroup
\end{table*}

\noindent\textbf{QA Generation.}
We generate QA pairs through four stages.
Prediction setup (S1) identifies task-specific targets and selects observation cutoffs using event timestamps and video inspection, ensuring that the queried event or result has not yet been revealed.
Offensive-player annotation (S2) assigns bounding boxes and frame-specific IDs to visible offensive players in each cutoff frame, adding player names when they can be reliably identified from the source annotations.
Predictive QA generation (S3) constructs prediction questions and their corresponding explanation questions.
Reference answers are determined by subsequent events, while supporting explanations use only evidence available at or before the cutoff.
Full-possession QA generation (S4) produces offensive and defensive reviews and outcome analyses, linking recommendations to observed execution and describing their potential benefits and risks.
Across both task families, the prompts require consideration of the remaining shot-clock time when it is readable.

\noindent\textbf{Question Filtering.}
We remove player-association questions when the queried action already determines its executor as the current ball handler, such as continuing a drive or holding the ball to survey the court.
These questions reduce to recognizing the observed ball handler and therefore provide limited assessment of predictive player association.

\noindent\textbf{Human Review.}
We use human review to correct errors in visual grounding and tactical interpretation introduced during model-assisted annotation.
Our review team comprises ten basketball enthusiasts with at least five years of experience following the sport and two basketball experts with competitive playing and coaching backgrounds, respectively.
All reviewers receive guideline training and jointly review examples to align their judgments.
The enthusiasts check whether questions and options are clear, answers are correct and supported by video evidence, and selected options agree with the answer content.
They also verify basketball terminology, the alignment of player boxes with IDs and names, and the completeness and formatting of annotations.
Errors are corrected in both QA pairs and player annotations.
Uncertain cases are discussed after re-examining the video evidence, with unresolved cases referred to the two experts for adjudication.
The experts assess action identification, tactical interpretation, and the appropriateness of recommendations.
Samples that remain unreliable after expert review are discarded.

\subsection{Benchmark Statistics}

BasketTactics contains 1,968 offensive possessions from 45 games, with 39,954 QA pairs retained after filtering and review.
Predictive tasks account for 34,050 QA pairs (85.22\%), comprising 10,468 for action prediction, 11,794 for tactic prediction, and 11,788 for outcome prediction.
Retrospective tasks contribute the remaining 5,904 QA pairs (14.78\%), with 1,968 each for offensive review, defensive review, and outcome analysis and adjustments.

We partition the dataset by game into training, validation, and test sets containing 1,377, 197, and 394 possessions, respectively.
The corresponding QA counts are 27,943, 3,995, and 8,016.
All possessions and associated QA pairs from each game remain within one split, ensuring no game overlap across splits.

Table~\ref{tab:benchmark_comparison} compares the scale and task coverage of sports QA benchmarks.
BasketTactics combines player-linked prediction and supporting explanations with full-possession offensive and defensive reviews.
It explicitly distinguishes tactical objective attainment from scoring outcomes and supports diagnosis and adjustment recommendations.

\begin{table*}[t]
\centering
\caption{
Predictive performance on BasketTactics.
Q1-Q3 assess future events, their results, and participating players, respectively.
Q1 reports accuracy, while Q2-Q3 report cumulative / conditional accuracy (\%).
Q4-Q6 (shaded) report conditional mean / coverage-weighted explanation scores, with a maximum of 15.
Bold indicates the best result for each metric.
}
\label{tab:predictive_results}

\begingroup
\newcommand{\predacc}[2]{#1\,/\,#2}
\scriptsize
\setlength{\tabcolsep}{2pt}
\renewcommand{\arraystretch}{1.15}

\resizebox{\textwidth}{!}{%
\begin{tabular}{
@{}l
*{3}{
ccc
>{\columncolor{black!4}}c
>{\columncolor{black!4}}c
>{\columncolor{black!4}}c
}
@{}
}
\toprule
\multirow{3}{*}{Model}
& \multicolumn{6}{c}{Action Prediction}
& \multicolumn{6}{c}{Tactic Prediction}
& \multicolumn{6}{c}{Outcome Prediction} \\
\cmidrule(lr){2-7}
\cmidrule(lr){8-13}
\cmidrule(lr){14-19}
& \multicolumn{3}{c}{Objective (Acc.\ \%)}
& \multicolumn{3}{>{\columncolor{black!8}}c}{Subjective (Score)}
& \multicolumn{3}{c}{Objective (Acc.\ \%)}
& \multicolumn{3}{>{\columncolor{black!8}}c}{Subjective (Score)}
& \multicolumn{3}{c}{Objective (Acc.\ \%)}
& \multicolumn{3}{>{\columncolor{black!8}}c}{Subjective (Score)} \\
\cmidrule(lr){2-4}
\cmidrule(lr){5-7}
\cmidrule(lr){8-10}
\cmidrule(lr){11-13}
\cmidrule(lr){14-16}
\cmidrule(lr){17-19}
& Q1 & Q2 & Q3 & Q4 & Q5 & Q6
& Q1 & Q2 & Q3 & Q4 & Q5 & Q6
& Q1 & Q2 & Q3 & Q4 & Q5 & Q6 \\
\midrule

\multicolumn{19}{c}{Commercial Models} \\
\midrule

GPT-5.6-Luna
& 52.28 & \predacc{49.75}{95.15} & \predacc{26.79}{67.62}
& \predacc{\textbf{14.02}}{7.33} & \predacc{\textbf{14.11}}{7.02} & \predacc{11.59}{3.11}
& 63.71 & \predacc{\textbf{60.41}}{\textbf{94.82}} & \predacc{38.07}{63.03}
& \predacc{\textbf{14.04}}{\textbf{8.94}} & \predacc{\textbf{14.08}}{\textbf{8.51}} & \predacc{11.57}{4.40}
& 52.79 & \predacc{33.25}{62.98} & \predacc{26.90}{80.92}
& \predacc{\textbf{13.65}}{7.21} & \predacc{\textbf{13.28}}{4.42} & \predacc{11.31}{3.04} \\

Gemini-3.8-Flash
& 50.51 & \predacc{46.45}{91.96} & \predacc{23.02}{63.54}
& \predacc{13.70}{6.92} & \predacc{13.71}{6.37} & \predacc{\textbf{12.51}}{2.88}
& 58.88 & \predacc{52.28}{88.79} & \predacc{36.29}{69.42}
& \predacc{13.53}{7.96} & \predacc{13.54}{7.08} & \predacc{\textbf{11.62}}{4.22}
& \textbf{57.11} & \predacc{\textbf{36.04}}{63.11} & \predacc{\textbf{29.70}}{82.39}
& \predacc{13.18}{\textbf{7.53}} & \predacc{12.94}{\textbf{4.66}} & \predacc{\textbf{12.17}}{\textbf{3.61}} \\

Claude Sonnet 5
& 43.65 & \predacc{41.37}{94.77} & \predacc{14.72}{47.56}
& \predacc{13.03}{5.69} & \predacc{13.40}{5.54} & \predacc{9.49}{1.40}
& 55.33 & \predacc{52.03}{94.04} & \predacc{25.13}{48.29}
& \predacc{13.17}{7.29} & \predacc{13.17}{6.85} & \predacc{10.89}{2.74}
& 49.49 & \predacc{31.22}{63.08} & \predacc{23.35}{74.80}
& \predacc{12.77}{6.32} & \predacc{12.52}{3.91} & \predacc{9.01}{2.10} \\

Qwen3.7-Plus
& 51.78 & \predacc{48.48}{93.63} & \predacc{15.47}{50.00}
& \predacc{13.10}{6.78} & \predacc{13.27}{6.43} & \predacc{10.27}{1.59}
& 56.85 & \predacc{50.51}{88.84} & \predacc{30.71}{60.80}
& \predacc{12.86}{7.31} & \predacc{13.03}{6.58} & \predacc{10.67}{3.28}
& 42.39 & \predacc{27.16}{64.07} & \predacc{20.30}{74.77}
& \predacc{12.76}{5.41} & \predacc{12.00}{3.26} & \predacc{10.59}{2.15} \\

\midrule
\multicolumn{19}{c}{Open-Source Models} \\
\midrule

InternVL3.5-8B
& 48.73 & \predacc{46.70}{\textbf{95.83}} & \predacc{16.98}{42.45}
& \predacc{11.48}{5.60} & \predacc{12.34}{5.76} & \predacc{7.60}{1.29}
& 46.45 & \predacc{37.06}{79.78} & \predacc{15.48}{41.78}
& \predacc{10.98}{5.10} & \predacc{11.40}{4.23} & \predacc{6.34}{0.98}
& 29.44 & \predacc{22.08}{75.00} & \predacc{13.96}{63.22}
& \predacc{11.88}{3.50} & \predacc{11.10}{2.45} & \predacc{6.20}{0.87} \\

Video-R2
& 49.75 & \predacc{38.83}{78.06} & \predacc{12.08}{35.56}
& \predacc{10.77}{5.36} & \predacc{11.24}{4.36} & \predacc{6.38}{0.77}
& 50.51 & \predacc{42.89}{84.92} & \predacc{9.39}{21.89}
& \predacc{9.45}{4.77} & \predacc{9.49}{4.07} & \predacc{6.78}{0.64}
& 25.13 & \predacc{11.42}{45.45} & \predacc{2.54}{22.22}
& \predacc{9.38}{2.36} & \predacc{7.40}{0.85} & \predacc{6.70}{0.17} \\

GLM-4.6V-Flash
& 51.78 & \predacc{46.95}{90.69} & \predacc{14.34}{29.69}
& \predacc{11.59}{6.00} & \predacc{11.71}{5.50} & \predacc{8.79}{1.26}
& 32.99 & \predacc{28.17}{85.38} & \predacc{10.66}{37.84}
& \predacc{11.10}{3.66} & \predacc{10.90}{3.07} & \predacc{8.10}{0.86}
& 17.26 & \predacc{8.63}{50.00} & \predacc{5.33}{61.76}
& \predacc{10.16}{1.75} & \predacc{8.62}{0.74} & \predacc{7.00}{0.37} \\

VideoLLaMA3-7B
& 36.80 & \predacc{24.11}{65.52} & \predacc{1.13}{7.32}
& \predacc{9.96}{3.66} & \predacc{10.35}{2.49} & \predacc{5.67}{0.06}
& 35.28 & \predacc{10.15}{28.78} & \predacc{2.54}{25.00}
& \predacc{8.91}{3.14} & \predacc{9.97}{1.01} & \predacc{5.40}{0.14}
& 15.99 & \predacc{3.05}{19.05} & \predacc{0.51}{16.67}
& \predacc{8.73}{1.40} & \predacc{7.42}{0.23} & \predacc{5.50}{0.03} \\

Qwen3-VL-8B
& 38.58 & \predacc{34.52}{89.47} & \predacc{7.17}{28.79}
& \predacc{11.29}{4.36} & \predacc{12.51}{4.32} & \predacc{7.53}{0.54}
& 44.67 & \predacc{37.56}{84.09} & \predacc{17.01}{45.27}
& \predacc{10.94}{4.89} & \predacc{11.57}{4.35} & \predacc{8.52}{1.45}
& 40.61 & \predacc{31.22}{\textbf{76.88}} & \predacc{22.59}{72.36}
& \predacc{11.62}{4.72} & \predacc{9.79}{3.06} & \predacc{7.93}{1.79} \\

Qwen3.5-9B
& 50.76 & \predacc{45.18}{89.00} & \predacc{14.34}{37.62}
& \predacc{11.31}{5.74} & \predacc{11.87}{5.36} & \predacc{8.00}{1.15}
& 52.79 & \predacc{41.62}{78.85} & \predacc{21.57}{51.83}
& \predacc{10.95}{5.78} & \predacc{11.40}{4.74} & \predacc{8.06}{1.74}
& 30.46 & \predacc{19.80}{65.00} & \predacc{12.94}{65.38}
& \predacc{10.89}{3.32} & \predacc{10.26}{2.03} & \predacc{8.22}{1.06} \\

Qwen3.5-9B (SFT)
& \textbf{62.44} & \predacc{\textbf{56.60}}{90.65} & \predacc{\textbf{44.91}}{\textbf{84.40}}
& \predacc{13.43}{\textbf{8.39}} & \predacc{13.49}{\textbf{7.63}} & \predacc{11.23}{\textbf{5.04}}
& \textbf{66.75} & \predacc{52.28}{78.33} & \predacc{\textbf{42.89}}{\textbf{82.04}}
& \predacc{13.16}{8.79} & \predacc{13.39}{7.00} & \predacc{10.91}{\textbf{4.68}}
& 52.79 & \predacc{31.22}{59.13} & \predacc{26.40}{\textbf{84.55}}
& \predacc{13.13}{6.93} & \predacc{13.16}{4.11} & \predacc{9.29}{2.45} \\

\bottomrule
\end{tabular}%
}
\endgroup
\end{table*}

\begin{table}[t]
\centering
\caption{Retrospective performance on BasketTactics.
The three retrospective task scores have a maximum of 20. Execution-Score is a supplementary six-class accuracy (\%) from offensive reviews.
}
\label{tab:retrospective_results}
\begingroup
\scriptsize
\setlength{\tabcolsep}{3pt}
\renewcommand{\arraystretch}{1.15}
\resizebox{\columnwidth}{!}{%
\begin{tabular}{@{}lcccc@{}}
\toprule
Model
& \shortstack{Offensive\\Review}
& \shortstack{Defensive\\Review}
& \shortstack{Outcome\\Analysis}
& \shortstack{Execution-\\Score} \\
\midrule

\multicolumn{5}{c}{Commercial Models} \\
\midrule
GPT-5.6-Luna     & \textbf{11.41} & \textbf{13.14} & \textbf{12.81} & 36.80 \\
Gemini-3.8-Flash & 11.12 & 11.90 & 12.74 & \textbf{40.86} \\
Claude Sonnet 5  &  9.14 &  9.93 &  9.84 & 35.03 \\
Qwen3.7-Plus     &  9.03 & 10.30 &  9.82 & 35.03 \\

\midrule
\multicolumn{5}{c}{Open-Source Models} \\
\midrule
InternVL3.5-8B   &  7.85 &  8.85 &  7.49 & 31.47 \\
Video-R2         &  6.52 &  6.41 &  7.20 & 26.90 \\
GLM-4.6V-Flash   &  7.57 &  8.53 &  7.42 & 28.17 \\
VideoLLaMA3-7B   &  7.07 &  7.24 &  7.22 & 19.54 \\
Qwen3-VL-8B      &  7.22 &  8.07 &  7.85 & 24.37 \\
Qwen3.5-9B       &  8.21 &  8.45 &  7.73 & 26.65 \\
Qwen3.5-9B (SFT)  &  8.24 & 11.27 & 10.30 & 27.16 \\

\bottomrule
\end{tabular}%
}
\endgroup
\end{table}

\subsection{Evaluation Protocol}
\label{sec:evaluation_protocol}

\noindent\textbf{Predictive Evaluation.}
Each prediction group follows the sequence Q1, Q4, Q2, Q5, Q3, and Q6.
A correct objective answer triggers its corresponding explanation question before evaluation proceeds to the next objective question.
An incorrect objective answer terminates the group.
Q1 and Q2 are evaluated against the reference options, while Q3 requires an exact match between the predicted and reference player-ID sets.
Player names are evaluated separately as an auxiliary accuracy metric.
Only ID correctness determines whether evaluation proceeds from Q3 to Q6.

\noindent\textbf{Explanation and Review Evaluation.}
GPT-6.1 Sol~\cite{gpt61} and Claude Opus 5-5~\cite{claudeOpus5} independently evaluate prediction explanations and retrospective answers against task-specific reference answers using shared rubrics.
Prediction explanations are scored on three dimensions comprising (i) factual consistency with observations available at or before the cutoff, (ii) evidential support for the prediction, and (iii) conceptual accuracy with clear use of basketball terminology.
Offensive and defensive reviews are scored on four dimensions comprising (i) identification of the main tactics or strategies together with the observed outcome, (ii) factual accuracy of the sequence of play, (iii) coverage of key mechanisms with accurate explanations, and (iv) conceptual accuracy with clear use of basketball terminology.
For outcome analysis and adjustments, the four dimensions comprise (i) diagnostic accuracy for the side under review, (ii) factual support for the analysis, (iii) relevant, feasible recommendations with an assessment of potential benefits and risks, and (iv) conceptual accuracy with clear use of basketball terminology.

Each judge assigns an integer score from 1 to 5 to each dimension.
These scores are summed separately for each judge, yielding totals of 3-15 for prediction explanations and 4-20 for retrospective answers.
The final score for each answer is the arithmetic mean of the total scores assigned by the two judges.
For each retrospective task, we report the mean score across its answers.
Prediction explanation scores are aggregated using the conditional and coverage-weighted measures defined below.
An analysis of agreement between human raters and the LLM judges is provided in the supplementary material.

\noindent\textbf{Score Aggregation.}
For objective questions, we report conditional accuracy over reached questions and cumulative accuracy over all valid questions.
For explanations, we report the conditional mean over reached questions and a coverage-weighted score computed as the sum of scores divided by the total number of valid questions.
Both cumulative accuracy and coverage-weighted scores assign zero to valid questions skipped after an earlier error.
All metrics are computed separately for each question type, excluding questions removed during dataset construction.
These metrics capture both performance on reached questions and the effect of early termination.

\noindent\textbf{Execution-Score.}
As a supplementary diagnostic metric, Execution-Score evaluates two judgments extracted from each offensive review.
These concern whether the final offensive phase actually uses an advantage created through teammate coordination and whether the possession results in scoring, no scoring, or drawing a defensive foul.
Their combinations yield six classes, for which we report classification accuracy against reference annotations.

\section{Experiment}

\subsection{Experimental Setup}

\begin{table*}[t]
\centering
\caption{
Predictive performance on BasketTactics with factual context from PL-NBA~\cite{pl-nba} restricted to the observation cutoff.
}
\label{tab:predictive_structured_inputs}

\begingroup
\newcommand{\predacc}[2]{#1\,/\,#2}
\newcommand{\pendingpair}{---\,/\,---}
\scriptsize
\setlength{\tabcolsep}{2pt}
\renewcommand{\arraystretch}{1.15}

\resizebox{\textwidth}{!}{%
\begin{tabular}{
@{}l
*{3}{
ccc
>{\columncolor{black!4}}c
>{\columncolor{black!4}}c
>{\columncolor{black!4}}c
}
@{}
}
\toprule
\multirow{3}{*}{Model}
& \multicolumn{6}{c}{Action Prediction}
& \multicolumn{6}{c}{Tactic Prediction}
& \multicolumn{6}{c}{Outcome Prediction} \\
\cmidrule(lr){2-7}
\cmidrule(lr){8-13}
\cmidrule(lr){14-19}
& \multicolumn{3}{c}{Objective (Acc.\ \%)}
& \multicolumn{3}{>{\columncolor{black!8}}c}{Subjective (Score)}
& \multicolumn{3}{c}{Objective (Acc.\ \%)}
& \multicolumn{3}{>{\columncolor{black!8}}c}{Subjective (Score)}
& \multicolumn{3}{c}{Objective (Acc.\ \%)}
& \multicolumn{3}{>{\columncolor{black!8}}c}{Subjective (Score)} \\
\cmidrule(lr){2-4}
\cmidrule(lr){5-7}
\cmidrule(lr){8-10}
\cmidrule(lr){11-13}
\cmidrule(lr){14-16}
\cmidrule(lr){17-19}
& Q1 & Q2 & Q3 & Q4 & Q5 & Q6
& Q1 & Q2 & Q3 & Q4 & Q5 & Q6
& Q1 & Q2 & Q3 & Q4 & Q5 & Q6 \\
\midrule

\multicolumn{19}{c}{Commercial Models} \\
\midrule

GPT-5.6-Luna
& 55.58 & \predacc{55.08}{\textbf{99.09}} & \predacc{27.55}{58.87}
& \predacc{\textbf{14.34}}{7.97} & \predacc{\textbf{14.24}}{7.84} & \predacc{13.75}{3.79}
& 66.24 & \predacc{\textbf{62.94}}{\textbf{95.02}} & \predacc{43.65}{69.35}
& \predacc{\textbf{14.16}}{\textbf{9.38}} & \predacc{13.67}{\textbf{8.61}} & \predacc{\textbf{14.02}}{6.12}
& 61.17 & \predacc{\textbf{38.83}}{63.49} & \predacc{33.50}{86.27}
& \predacc{\textbf{14.29}}{\textbf{8.74}} & \predacc{13.56}{\textbf{5.27}} & \predacc{{13.92}}{{4.66}} \\

Gemini-3.8-Flash
& 58.38 & \predacc{52.03}{89.13} & \predacc{32.08}{74.56}
& \predacc{13.85}{8.08} & \predacc{13.82}{7.19} & \predacc{\textbf{14.41}}{4.62}
& 65.48 & \predacc{58.88}{89.92} & \predacc{44.42}{75.43}
& \predacc{13.54}{8.86} & \predacc{13.39}{7.88} & \predacc{13.86}{6.15}
& \textbf{63.20} & \predacc{37.31}{59.04} & \predacc{\textbf{33.76}}{\textbf{90.48}}
& \predacc{13.45}{8.50} & \predacc{12.80}{4.78} & \predacc{\textbf{14.07}}{\textbf{4.75}} \\

Claude Sonnet 5
& 52.79 & \predacc{50.00}{94.71} & \predacc{15.85}{39.25}
& \predacc{13.89}{7.34} & \predacc{13.84}{6.92} & \predacc{12.90}{2.05}
& 59.39 & \predacc{55.58}{93.59} & \predacc{28.43}{51.14}
& \predacc{13.50}{8.02} & \predacc{13.32}{7.40} & \predacc{13.35}{3.79}
& 61.93 & \predacc{36.29}{58.61} & \predacc{27.92}{76.92}
& \predacc{13.73}{8.50} & \predacc{13.16}{4.78} & \predacc{13.57}{3.79} \\

Qwen3.7-Plus
& 55.08 & \predacc{53.30}{96.77} & \predacc{20.00}{53.00}
& \predacc{13.85}{7.63} & \predacc{13.61}{7.26} & \predacc{13.42}{2.68}
& 64.47 & \predacc{57.36}{88.98} & \predacc{33.76}{58.85}
& \predacc{13.45}{8.67} & \predacc{12.59}{7.22} & \predacc{13.41}{4.53}
& 48.98 & \predacc{30.96}{63.21} & \predacc{23.86}{77.05}
& \predacc{13.66}{6.69} & \predacc{12.80}{3.96} & \predacc{13.44}{3.21} \\

\midrule
\multicolumn{19}{c}{Open-Source Models} \\
\midrule

Qwen3.5-9B
& 55.08 & \predacc{51.27}{93.09} & \predacc{15.47}{37.61}
& \predacc{11.35}{6.25} & \predacc{11.50}{5.90} & \predacc{10.98}{1.70}
& 56.85 & \predacc{45.69}{80.36} & \predacc{22.59}{49.44}
& \predacc{10.81}{6.14} & \predacc{10.87}{4.96} & \predacc{10.78}{2.43}
& 43.65 & \predacc{29.70}{\textbf{68.02}} & \predacc{19.80}{66.67}
& \predacc{10.70}{4.67} & \predacc{10.36}{3.08} & \predacc{11.79}{2.33} \\

Qwen3.5-9B (SFT)
& \textbf{62.94} & \predacc{\textbf{58.63}}{93.15} & \predacc{\textbf{42.26}}{\textbf{81.75}}
& \predacc{13.59}{\textbf{8.55}} & \predacc{13.68}{\textbf{8.02}} & \predacc{{13.80}}{\textbf{5.83}}
& \textbf{68.27} & \predacc{55.33}{81.04} & \predacc{\textbf{44.92}}{\textbf{81.19}}
& \predacc{13.15}{8.98} & \predacc{\textbf{13.69}}{7.57} & \predacc{13.75}{\textbf{6.18}}
& 59.14 & \predacc{35.79}{60.52} & \predacc{30.96}{86.52}
& \predacc{13.73}{8.12} & \predacc{\textbf{13.64}}{4.88} & \predacc{13.66}{4.23} \\

\bottomrule
\end{tabular}%
}
\endgroup
\end{table*}

\begin{table}[t]
\centering
\caption{
Retrospective performance on BasketTactics with factual context from PL-NBA~\cite{pl-nba}.
The context covers complete possessions, including recorded action outcomes.
}
\label{tab:retrospective_structured_inputs}

\begingroup
\footnotesize
\setlength{\tabcolsep}{3pt}
\renewcommand{\arraystretch}{1.1}

\begin{tabular*}{\columnwidth}
{@{\extracolsep{\fill}}lcccc@{}}
\toprule
Model
& \shortstack{Offensive\\Review}
& \shortstack{Defensive\\Review}
& \shortstack{Outcome\\Analysis}
& \shortstack{Execution-\\Score}\\
\midrule

\multicolumn{5}{c}{Commercial Models} \\
\midrule
GPT-5.6-Luna     & \textbf{17.99} & \textbf{17.77} & \textbf{16.01} & 69.29 \\
Gemini-3.8-Flash & 16.98 & 16.07 & 15.79 & \textbf{76.65} \\
Claude Sonnet 5  & 16.49 & 15.20 & 14.53 & 74.11 \\
Qwen3.7-Plus     & 16.28 & 14.75 & 13.80 & 72.84 \\

\midrule
\multicolumn{5}{c}{Open-Source Models} \\
\midrule
Qwen3.5-9B       & 12.79 & 10.26 &  8.81 & 62.69 \\
Qwen3.5-9B (SFT)  & 15.73 & 15.04 & 13.11 & 64.72 \\

\bottomrule
\end{tabular*}
\endgroup
\end{table}

\noindent\textbf{Baseline models.}
We evaluate ten MLLMs on BasketTactics in a zero-shot setting using the same task-specific prompt templates. Commercial models include GPT-5.6-Luna~\cite{gpt56luna}, Gemini-3.8-Flash~\cite{gemini38flash}, Claude Sonnet 5~\cite{claudesonnet5}, and Qwen3.7-Plus~\cite{qwen37plus}.
Open-source models include InternVL3.5-8B~\cite{internvl35}, Video-R2~\cite{videor2}, GLM-4.6V-Flash~\cite{glm46v}, VideoLLaMA3-7B~\cite{videollama3}, Qwen3-VL-8B~\cite{qwen3vl}, and Qwen3.5-9B~\cite{qwen35}.
We additionally evaluate a supervised fine-tuned version of Qwen3.5-9B~\cite{qwen35} to examine the effect of task-specific training on prediction and retrospective analysis.

\noindent\textbf{Implementation Details.}
Videos are sampled at 2 frames per second, including the cutoff frame for each prediction prefix and the final frame for each full-possession review.
Within each prediction group, subsequent questions retain earlier objective questions and model answers but exclude generated explanations.
Each prediction group and each retrospective question is evaluated in a separate context.

We fine-tune Qwen3.5-9B~\cite{qwen35} on the training split using LoRA~\cite{lora} in the language backbone while updating the visual merger.
Training covers predictive tasks, their explanations, and retrospective tasks.
For multiple-choice training questions, answer options are removed from the prompts, with the correct option text used as the supervision target.
The judges in Section~\ref{sec:evaluation_protocol} receive text-only inputs comprising the question, model response, and reference answer.
For prediction explanations, they receive the corresponding objective question and answer, together with the observation cutoff.
Complete implementation settings, judging prompts, and auxiliary player-name accuracy results are provided in the supplementary material.

\subsection{Main Results}
\label{sec:main_results}

\noindent\textbf{Predictive Performance.}
Table~\ref{tab:predictive_results} shows that jointly predicting future events, their results, and the participating players remains challenging.
The best zero-shot cumulative Q3 accuracies are 26.79\%, 38.07\%, and 29.70\% for action, tactic, and outcome prediction, respectively.
These accuracies reflect both preceding prediction errors and errors in player association.
Explanation scores further illustrate the effect of early termination.
For action-related player explanations (Q6), GPT-5.6-Luna obtains a conditional mean of 11.59 but a coverage-weighted score of 3.11, reflecting the limited coverage of correct prediction sequences.
Fine-tuning Qwen3.5-9B improves Q1 accuracy and cumulative Q3 accuracy across all three prediction groups.
The fine-tuned model achieves Q1 accuracies of 62.44\% and 66.75\% for action and tactic prediction, respectively, surpassing all evaluated zero-shot models.
However, cumulative Q3 accuracy remains below 45\% for every model across all three groups, indicating persistent difficulty in correctly predicting events, their results, and the associated players within the same sequence.

\noindent\textbf{Retrospective Performance.}
Table~\ref{tab:retrospective_results} shows that commercial models outperform zero-shot open-source models, but their rankings differ across evaluation criteria.
GPT-5.6-Luna leads all three retrospective scores, whereas Gemini-3.8-Flash achieves the highest Execution-Score of 40.86\%.
This ranking reversal suggests that Execution-Score provides a complementary assessment of explicit execution and outcome judgments beyond the overall offensive review score.
Fine-tuning Qwen3.5-9B improves defensive review and outcome analysis by 2.82 and 2.57 points, respectively, compared with only 0.03 points in offensive review.
Its Execution-Score increases by just 0.51\%.
The gains from this fine-tuning setup are therefore concentrated in defensive review and outcome analysis, with only small changes in offensive review and Execution-Score.

\begin{figure*}[t]
    \centering
    \includegraphics[width=\textwidth]{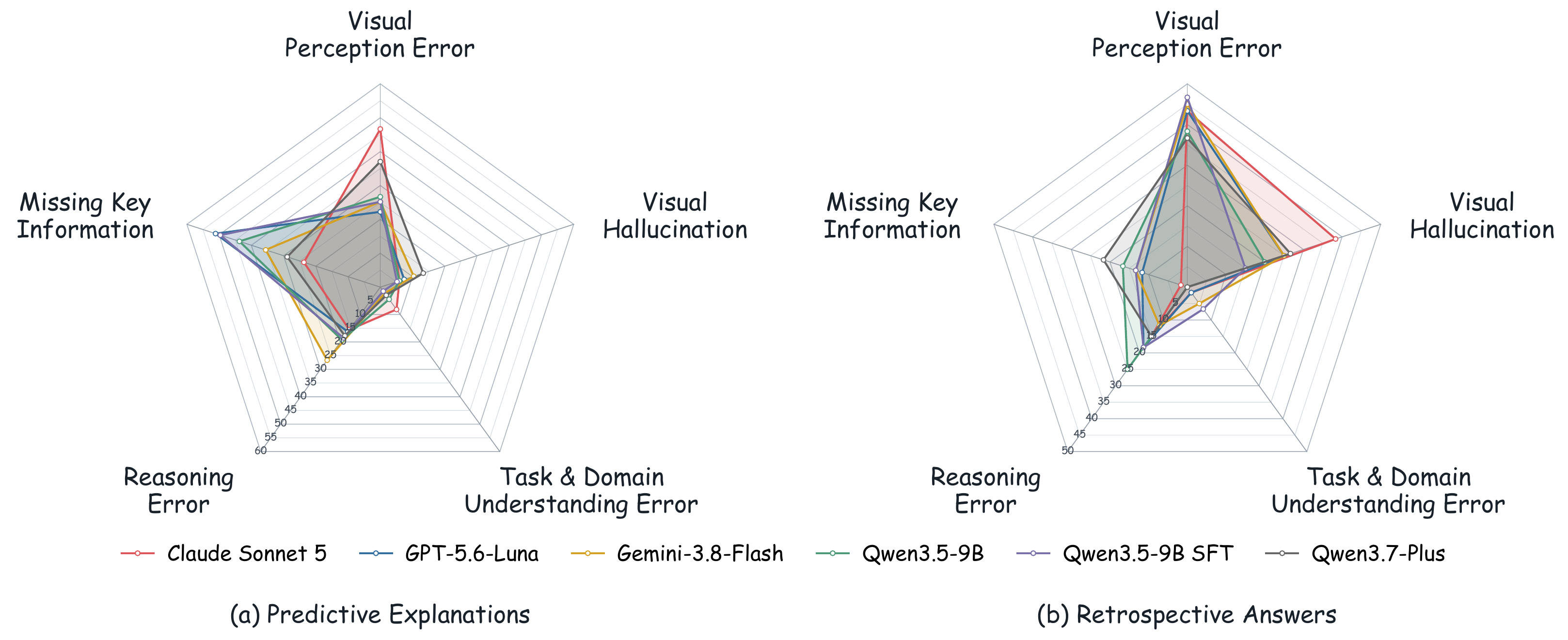}
    \caption{
    Distribution of primary error types across model configurations for (a) predictive explanations and (b) retrospective answers. Responses are manually annotated according to their primary error category. Missing key information is prevalent in predictive explanations, whereas visual perception errors dominate retrospective answers.
        }
    \label{fig:error_analysis}
\end{figure*}

\subsection{Effect of Factual Context}
\label{sec:factual_context}

Tactical analysis requires models to interpret player interactions from basic facts about who acts, when, and where. We examine whether explicitly providing these facts supports prediction and retrospective analysis.

\noindent\textbf{Experimental Setting.}
We supplement videos with chronological action records from PL-NBA~\cite{pl-nba}, specifying player names, basic actions, time intervals, and positions or movements.
Prediction records contain only information available by the observation cutoff, with ongoing actions truncated at that point.
Retrospective records cover the complete possession, including action outcomes.
No explicit tactical or coordination annotations are provided.
We evaluate four commercial models alongside the zero-shot and fine-tuned versions of Qwen3.5-9B under this setting.

\noindent\textbf{Results and Analysis.}
Tables~\ref{tab:predictive_structured_inputs} and~\ref{tab:retrospective_structured_inputs} report results with factual context, showing higher Q1 accuracies and retrospective scores across all evaluated configurations than those in Tables~\ref{tab:predictive_results} and~\ref{tab:retrospective_results}.
For zero-shot Qwen3.5-9B, outcome Q1 accuracy rises from 30.46\% to 43.65\%, while offensive review improves from 8.21 to 12.79.
These changes suggest that explicit action records support both anticipation of subsequent events and interpretation of observed play.
Offensive review shows the largest retrospective gain for every configuration, consistent with the value of temporal action records for reconstructing offensive execution.
However, cumulative Q3 accuracy remains below 45\% across all prediction groups, indicating that maintaining correct predictions through the participant question remains challenging even with basic action facts.

\subsection{Error Analysis}
\label{sec:error_analysis}

To understand the errors underlying model performance, we analyze 810 prediction explanations and 360 retrospective answers sampled from six model configurations.
We manually categorize the primary error in each response as a visual perception error, visual hallucination, task and domain understanding error, reasoning error, or missing key information.
Detailed sampling procedures and annotation criteria are provided in the supplementary material.

Figure~\ref{fig:error_analysis} shows distinct error profiles in the sampled responses.
Missing key information is the largest category for four configurations in prediction explanations.
This suggests that models often provide plausible rationales without covering the evidence needed to fully justify their predictions, making completeness an important limitation beyond factual correctness.
For retrospective reviews, visual perception errors dominate across all six configurations.
Access to the complete possession therefore does not ensure accurate reconstruction of player actions, spatial relations, or event sequences.
Claude Sonnet 5 has the highest proportion of visual hallucinations among the six configurations in the sampled retrospective responses, with this category ranking second only to visual perception errors within its error distribution.
The supplementary material provides qualitative visualizations of successful and failed cases, illustrating representative model behaviors and error patterns.

\section{Conclusion}
\label{sec:conclusion}

We introduce BasketTactics, a benchmark for evaluating possession-level tactical reasoning in MLLMs.
It pairs prediction under explicit observation cutoffs with full-possession offensive and defensive reviews, distinguishing tactical objective attainment from realized scoring.
Our evaluation reveals difficulties in jointly predicting events and identifying their participants, providing complete supporting evidence, and reconstructing tactical execution.
Providing factual context improves initial event prediction and retrospective analysis, with the largest retrospective gains in offensive review.
These findings identify evidence use and execution reconstruction as important challenges for current models.
We hope BasketTactics advances basketball tactical reasoning in MLLMs, improving their ability to anticipate coordinated play, interpret offensive and defensive interactions, and assess tactical execution.

\section*{LLM Usage Statement}
We use GPT-6 Astra to assist QA generation, with human review and correction. GPT-6.1 Sol and Claude Opus 5-5 score explanations and retrospective answers, and Codex helps revise the manuscript.  Authors take responsibility for submitted content.
{
    \small
    \bibliographystyle{ieeenat_fullname}
    \bibliography{main}
}

\definecolor{RubricLine}{gray}{0.72}
\definecolor{RubricBackground}{gray}{0.975}
\definecolor{RubricHeading}{gray}{0.925}
\newtcolorbox{rubricpanel}[1]{%
  enhanced jigsaw, breakable, sharp corners,
  colframe=RubricLine, colback=RubricBackground,
  colbacktitle=RubricHeading, coltitle=black,
  boxrule=0.4pt, titlerule=0.4pt, boxsep=0pt,
  left=6pt, right=6pt, top=5pt, bottom=5pt,
  toptitle=4pt, bottomtitle=4pt,
  before skip=7pt plus 2pt, after skip=7pt plus 2pt,
  lines before break=4,
  fonttitle=\fontsize{10.5}{12}\selectfont\bfseries,
  fontupper=\fontsize{9}{10.7}\selectfont,
  before upper={\raggedright\setlength{\parskip}{0.8pt}},
  title={#1},
  title after break={#1\space(continued)}
}
\newenvironment{rubricdimension}[1]{%
  \par\addvspace{3pt}\noindent\textbf{#1}\par\nobreak
}{\par}
\newcommand{\promptmessage}[1]{\par\noindent\textbf{#1}\par\nobreak}
\newcommand{\rubricscore}[2]{%
  \par\noindent\hangindent=1.25em\hangafter=1\relax
  \makebox[1.25em][l]{\textbf{#1}}#2\par
}

\clearpage
\twocolumn[{\centering\Large\bfseries Supplementary Material\par\vspace{1em}}]
\appendix

\section{BasketTactics Benchmark}

\subsection{Data Availability Statement}

We plan to make the BasketTactics annotations and dataset splits publicly
available. Researchers may request access to the original broadcast clips
by email for non-commercial research. Requests will be reviewed
individually, and approved applicants must sign a research-use agreement
prohibiting redistribution before receiving access instructions.

\subsection{Observation Cutoff Selection}

For each prediction task, we select the observation cutoff according to the criteria below.

\begin{itemize}
\item \textbf{Action prediction.} We truncate the video before the target action is clearly visible. The video prefix retains cues such as ball control, player positions, and preceding actions.
\item \textbf{Tactic prediction.} We truncate the video before the target offensive play is clearly visible. The video prefix captures the offensive setup through cues such as player positions, directions of movement, and preparations for a screen.
\item \textbf{Outcome prediction.} We truncate the video before the possession-ending action or terminal event is clearly visible. The video prefix captures the preceding offensive sequence, including how the ball moves, as well as the positions of the ball handler and nearby defenders relative to the basket.
\end{itemize}

\subsection{Expert Review Examples}
\begin{figure}[t]
    \centering
    \includegraphics[width=\columnwidth]{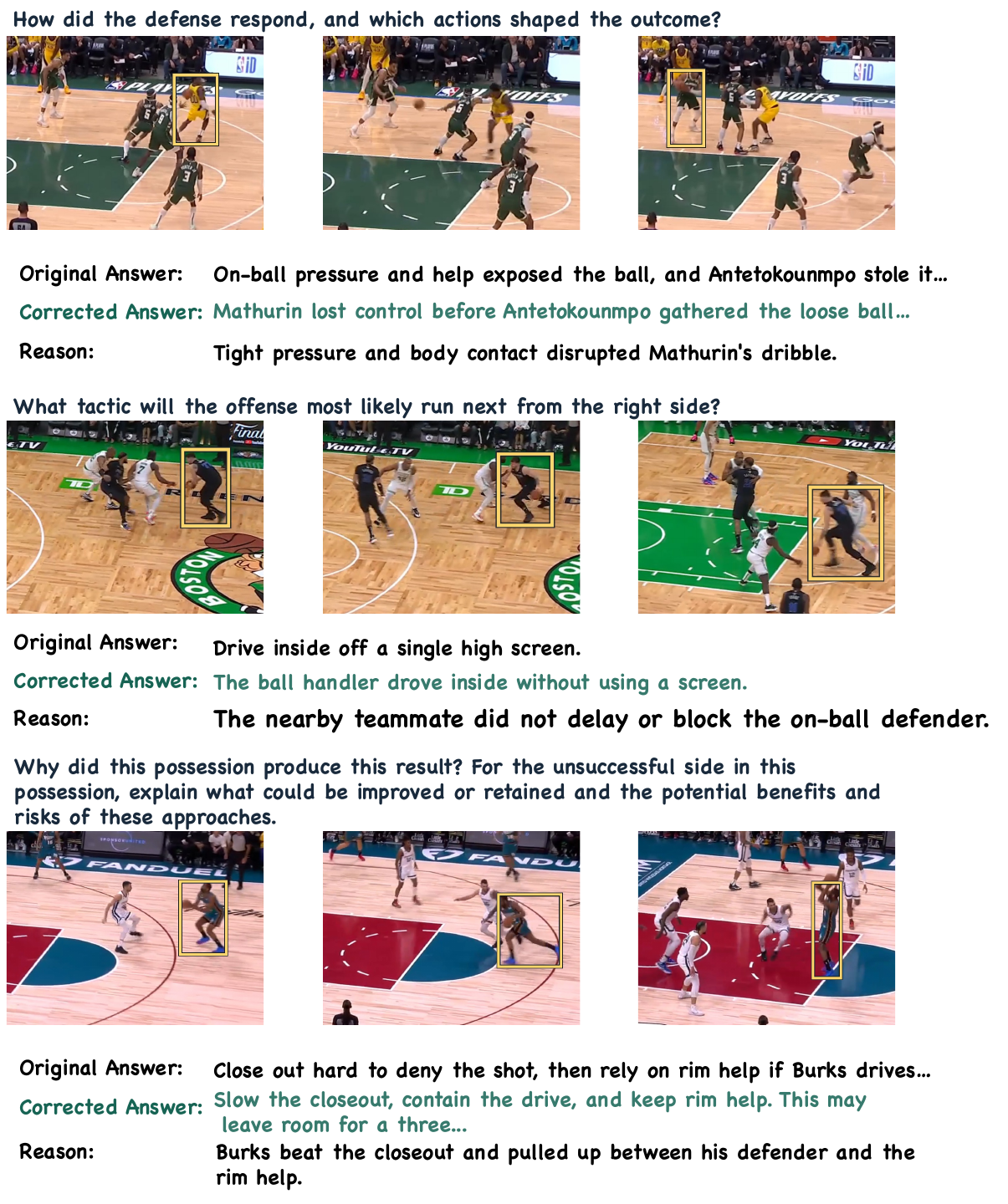}
    \caption{Expert review examples from BasketTactics. The first case illustrates
    action identification. The second case illustrates tactical interpretation. The third case focuses on the appropriateness of recommendations.}
    \label{fig:expert_review_examples}
\end{figure}

Figure~\ref{fig:expert_review_examples} illustrates the three aspects of expert review, namely action identification, tactical
interpretation, and the appropriateness of recommendations. Below each example the first row presents the initial answer. The second row
gives the corrected answer and the third row explains why it was revised. 

\subsection{Explanation and Review Evaluation}

\noindent\textbf{Scoring Dimensions.}
The rubrics below specify the criteria for each score level (1-5) within the dimensions introduced in the main paper. They cover prediction explanations (Q4-Q6), offensive and defensive reviews (R1-R2), and outcome analysis and adjustments (R3).
\begin{rubricpanel}{Prediction Explanations (Q4-Q6)}
\begin{rubricdimension}{D1. Factual Consistency}
\rubricscore{5}{Key facts are accurate and consistent with observations available at or before the cutoff.}
\rubricscore{4}{The central facts are correct. Minor inaccuracies do not affect the main rationale.}
\rubricscore{3}{The answer includes correct facts but makes some substantive errors or omits key facts.}
\rubricscore{2}{The answer makes clear errors about important facts, offensive or defensive interactions, or timing. Few relevant facts remain correct.}
\rubricscore{1}{The answer contradicts core facts, relies mainly on a future outcome, or has no valid factual basis.}
\end{rubricdimension}
\begin{rubricdimension}{D2. Evidence-Based Reasoning}
\rubricscore{5}{The answer uses a specific basketball mechanism to explain why the observed cues support the prediction.}
\rubricscore{4}{The main rationale and mechanism are complete, with minor gaps.}
\rubricscore{3}{The rationale is relevant, but the link between evidence and prediction is incomplete or partly generic.}
\rubricscore{2}{The answer mostly restates the prediction, gives a generic or weakly related reason, or explains a different prediction.}
\rubricscore{1}{The answer gives no valid reason, presents evidence that contradicts its conclusion, or makes a clear causal error.}
\end{rubricdimension}
\begin{rubricdimension}{D3. Basketball Concepts and Terminology}
\rubricscore{5}{Basketball concepts, court locations, actions, and player roles are accurate and described clearly.}
\rubricscore{4}{The answer uses basketball concepts accurately overall and has minor terminology or wording issues.}
\rubricscore{3}{The answer shows a generally correct understanding of basketball concepts but is ambiguous or repetitive.}
\rubricscore{2}{Misuse of key basketball concepts makes the explanation difficult to understand.}
\rubricscore{1}{The answer confuses basketball concepts and fails to provide a correct explanation.}
\end{rubricdimension}
\end{rubricpanel}

\begin{rubricpanel}{Offensive or Defensive Review (R1 or R2)}
\begin{rubricdimension}{D1. Tactic and Outcome Identification}
\rubricscore{5}{The answer identifies the main offensive or defensive tactic and accurately states the observed outcome.}
\rubricscore{4}{The main judgments are correct, with a broad tactical label or a minor omission.}
\rubricscore{3}{The offensive or defensive tactic or the observed outcome is materially inaccurate or missing.}
\rubricscore{2}{The answer clearly misidentifies the main offensive or defensive tactic and contains little relevant information.}
\rubricscore{1}{The central judgment is wrong, the answer reverses the offensive and defensive perspectives, or it gives no meaningful analysis.}
\end{rubricdimension}
\begin{rubricdimension}{D2. Factual Consistency}
\rubricscore{5}{The answer accurately describes key player roles, positions, the order of actions, and changes in possession.}
\rubricscore{4}{Minor errors in detail do not affect the main explanation.}
\rubricscore{3}{The answer mixes accurate details with significant errors or omissions. The main sequence remains recognizable.}
\rubricscore{2}{The answer misstates several important events, their order, or relationships between players.}
\rubricscore{1}{The answer gets most of the central sequence wrong, invents events, or provides no meaningful account.}
\end{rubricdimension}
\begin{rubricdimension}{D3. Tactical Accuracy and Completeness}
\rubricscore{5}{The answer covers key events, explains how the offense or defense changes the spacing, and connects the creation or loss of an opportunity to the observed outcome.}
\rubricscore{4}{The explanation of the main mechanism is sound, with a few minor links left unclear.}
\rubricscore{3}{The sequence is generally correct. The answer only partly explains the mechanism or omits a key event.}
\rubricscore{2}{The answer mostly lists actions or names a tactic and says little about how the offense or defense worked.}
\rubricscore{1}{The answer omits the key explanation or describes a mechanism that clearly contradicts the observed sequence.}
\end{rubricdimension}
\begin{rubricdimension}{D4. Basketball Concepts and Terminology}
\rubricscore{5}{Basketball concepts are accurate, the offensive or defensive perspective is clear, and the account is coherent.}
\rubricscore{4}{The answer has minor issues with terminology or organization, but the overall explanation remains clear.}
\rubricscore{3}{Some concepts or the perspective are unclear, but the answer remains understandable.}
\rubricscore{2}{Multiple errors in key concepts hinder understanding.}
\rubricscore{1}{Confused concepts and narration prevent a usable offensive or defensive review.}
\end{rubricdimension}
\end{rubricpanel}

\begin{rubricpanel}{Outcome Analysis and Adjustments (R3)}
\begin{rubricdimension}{D1. Diagnostic Accuracy}
\rubricscore{5}{The answer correctly identifies the unsuccessful side and the main reasons for the outcome. It distinguishes the effects of tactical choices, execution, the opponent's response, and individual ability.}
\rubricscore{4}{The main reasons are correct, but the explanation of secondary factors is incomplete.}
\rubricscore{3}{The diagnosis is plausible but generic or one-sided, or confuses the execution of the play with variability in the outcome.}
\rubricscore{2}{The stated cause has a weak connection to the outcome, or the diagnosis rests solely on whether a shot went in.}
\rubricscore{1}{The answer identifies the wrong side, gives a central diagnosis that contradicts the observed events, or provides no valid diagnosis.}
\end{rubricdimension}
\begin{rubricdimension}{D2. Evidence-Based Analysis}
\rubricscore{5}{The answer supports its diagnosis and recommendations with specific, accurate facts from the possession.}
\rubricscore{4}{The central facts are accurate, with minor inaccuracies.}
\rubricscore{3}{The answer includes relevant facts, but an important judgment is unsupported or key information is missing.}
\rubricscore{2}{Several key premises are wrong, or most of the analysis is disconnected from this possession.}
\rubricscore{1}{The analysis relies mainly on invented, contradictory, or unrelated facts.}
\end{rubricdimension}
\begin{rubricdimension}{D3. Recommendation Suitability and Trade-Offs}
\rubricscore{5}{The recommendation addresses the main issue and fits the spacing, timing, and the players. The response explains potential benefits and risks and justifies keeping the current strategy when applicable.}
\rubricscore{4}{The recommendation is specific and feasible and covers the main benefits and risks, with minor gaps.}
\rubricscore{3}{The recommendation is generally reasonable but has a clear gap in relevance, conditions for execution, benefits, or risks.}
\rubricscore{2}{The recommendation is generic, lacks actionable details, conflicts with on-court conditions, or assesses benefits and risks unevenly.}
\rubricscore{1}{The answer offers no useful recommendation, targets the wrong side, or proposes an infeasible and irrelevant adjustment.}
\end{rubricdimension}
\begin{rubricdimension}{D4. Basketball Concepts and Terminology}
\rubricscore{5}{Basketball concepts are accurate, and the answer clearly distinguishes observed facts from assumptions underlying a recommendation.}
\rubricscore{4}{The answer has minor terminology or wording issues.}
\rubricscore{3}{Some concepts or distinctions between facts and assumptions are unclear, or the answer is repetitive.}
\rubricscore{2}{The answer misuses a key basketball concept or describes a proposed change as an observed event.}
\rubricscore{1}{The presentation is too confused to identify a useful analysis or claim.}
\end{rubricdimension}
\end{rubricpanel}

\par\smallskip
\noindent\textbf{Human-LLM Agreement.}
For each of the six model configurations, we randomly sampled 50 prediction explanations and 50 retrospective answers, yielding 300 responses per task family. Human evaluators scored the sampled responses using the same rubrics as the LLM judges. For each response, the LLM-judge score was the arithmetic mean of the total scores assigned by Sol and Opus. Table~\ref{tab:supp-human-llm-agreement} compares the mean human and LLM-judge scores for each model configuration and task family.

\par\smallskip\noindent
\begin{minipage}{\columnwidth}
\captionsetup{type=table}
\caption{Mean human and LLM-judge scores on sampled prediction explanations and retrospective answers.}
\label{tab:supp-human-llm-agreement}
\centering
\footnotesize
\setlength{\tabcolsep}{4pt}
\begin{tabular}{@{}lrrrr@{}}
\toprule
& \multicolumn{2}{c}{Prediction (/15)} & \multicolumn{2}{c}{Review (/20)} \\
\cmidrule(lr){2-3}\cmidrule(lr){4-5}
Model configuration & Human & LLM judge & Human & LLM judge \\
\midrule
GPT-5.6-Luna        & 13.68 & 13.81 & 12.90 & 12.53 \\
Gemini-3.8-Flash    & 12.28 & 12.85 & 12.54 & 12.35 \\
Claude Sonnet 5     & 11.94 & 12.59 &  8.86 &  9.60 \\
Qwen3.7-Plus        & 12.10 & 12.79 &  8.88 &  8.85 \\
Qwen3.5-9B          &  9.96 & 10.60 &  7.94 &  8.06 \\
Qwen3.5-9B (SFT)    & 13.54 & 12.97 & 10.60 & 10.23 \\
\midrule
\textbf{Overall} & \textbf{12.25} & \textbf{12.60} & \textbf{10.29} & \textbf{10.27} \\
\bottomrule
\end{tabular}
\end{minipage}\par

\section{Experiment}

\subsection{Experimental Setup }

\noindent\textbf{System Prompt.}
We present two system prompt templates for the evaluated models, one shared by all prediction questions (Q1-Q6) and one shared by all retrospective questions (R1-R3).

\begin{rubricpanel}{Predictive Tasks (Q1-Q6)}
You are a basketball game analyst.

Base predictions only on actions, court positions, movement, passing lanes, defense, spacing, and readable clocks already visible in the video provided. For rationale questions, explain the specific reasons for the prediction you just made.

Answer only the target of the current question. Describe positions using court regions such as the top of the arc, wing, corner, free-throw line, low post, and paint. Distinguish holding the ball, dribbling, driving, changing direction, and making a spin move.
\end{rubricpanel}

\begin{rubricpanel}{Retrospective Tasks (R1-R3)}
You are a basketball game analyst. Answer the current full-possession review question.

Describe the actual offensive and defensive sequence and outcome based only on the video provided.

For the offensive review, explain how the coordination created an opportunity and what the final outcome was. For the defensive review, explain how the defense responded and which factors affected the outcome.

For the outcome analysis, offer specific recommendations only for the side or sides that were unsuccessful in this possession, and explain the benefits and risks of those recommendations. The offense is unsuccessful if it does not score. The defense is unsuccessful if it concedes points or commits a foul. Apply these criteria independently to each side.

Answer the current question directly, describing the actual sequence concretely. Begin the offensive review with “Offensive tactic [specific name].” Begin the defensive review with “Defensive strategy [specific name].” Then explain the actual sequence, mechanism, and outcome.
\end{rubricpanel}

\noindent\textbf{Supervised Fine-Tuning.}
We fine-tune Qwen3.5-9B by jointly training the LoRA adapters in the language backbone and the visual merger. Table~\ref{tab:supp_sft} lists the learning rates and other training hyperparameters. We compute the autoregressive cross-entropy loss only over the reference
answer tokens for the current question. Given the input $x$ and the reference answer $y = (y_1, \ldots, y_T)$, the loss at each answer position $t$ is
\begin{equation}
    \ell_t = -\log p_{\theta}\!\left(y_t \mid x, y_{<t}\right).
    \label{eq:sft_token_loss}
\end{equation}

\begin{table}[t]
\centering
\caption{Supervised fine-tuning configuration for Qwen3.5-9B.}
\label{tab:supp_sft}
\small
\begin{tabular}{@{}p{0.54\linewidth}@{\hspace{6pt}}p{\dimexpr0.46\linewidth-6pt\relax}@{}}
\toprule
\textbf{Parameter} & \textbf{Value} \\
\midrule
Trainable components & Language LoRA and visual merger \\
LoRA rank / alpha / dropout & 32 / 64 / 0.05 \\
Language LoRA learning rate & $2\times10^{-5}$ \\
Visual merger learning rate & $5\times10^{-6}$ \\
Weight decay / gradient clipping & 0.01 / 1.0 \\
Batch size & 32 \\
Training epochs & 3 \\
\bottomrule
\end{tabular}
\end{table}

\subsection{Quantitative Analysis}

\noindent\textbf{Auxiliary Player-Name Accuracy.}
Table~2 of the main paper evaluates player identification in Q3 using the box IDs assigned to players in the corresponding frame. For each model, we measure player-name accuracy on the subset of Q3 responses whose predicted box-ID set exactly matches the reference set. Across the three prediction tasks, Table~\ref{tab:supp_player_names} reports the percentage of these responses for which the full set of predicted player names exactly matches the
reference name set.
\begin{table}[t]
\centering
\caption{Player-name accuracy in Q3. For each model, we consider only responses with the correct set of player-box IDs. We report the percentage of these responses for which the predicted set of player names also exactly matches the reference set.}
\label{tab:supp_player_names}
\small
\begin{tabular}{@{}lr@{}}
\toprule
\textbf{Model} & \textbf{Name accuracy (\%)} \\
\midrule
GPT-5.6-Luna & 40.37 \\
Gemini-3.8-Flash & 55.76 \\
Claude Sonnet 5 & 29.57 \\
Qwen3.7-Plus & 33.88 \\
\midrule
InternVL3.5-8B & 8.07 \\
Video-R2 & 8.86 \\
GLM-4.6V-Flash & 10.89 \\
VideoLLaMA3-7B & 0.00 \\
Qwen3-VL-8B & 12.00 \\
Qwen3.5-9B & 16.09 \\
Qwen3.5-9B (SFT) & 27.30 \\
\bottomrule
\end{tabular}
\end{table}

\noindent\textbf{Effect of Factual Context.}
For the experiments in Section~4.3 of the main paper, we build action timelines from PL-NBA records and match the records to BasketTactics annotations to add court locations and player movements. For prediction tasks, the records cover only the observed video prefix and specify the player's name, action, time interval, and available location or
movement information for each completed action. For retrospective tasks, the records cover the full possession and include available action outcomes.
The panels illustrate tactic prediction (Q1) and offensive review (R1) for the same test possession.

\begin{rubricpanel}{Tactic prediction (Q1)}
\textbf{Observation cutoff:} 18.700~s.\par
\smallskip
\textbf{0.000\textendash{}0.833~s}\enspace Ryan Dunn inbounded the ball. Location / Movement: out of bounds behind the baseline beneath the backcourt basket.\par
\textbf{1.000\textendash{}7.000~s}\enspace Tyus Jones dribbled the ball. Location / Movement: advancing to the top of the arc in the frontcourt.\par
\textbf{7.000\textendash{}8.000~s}\enspace Tyus Jones passed the ball. Location / Movement: beyond the three-point line at the top of the arc.\par
\textbf{8.000\textendash{}9.667~s}\enspace Devin Booker dribbled the ball. Location / Movement: driving inward after receiving the ball on the perimeter.\par
\textbf{10.167\textendash{}11.000~s}\enspace Devin Booker passed the ball. Location / Movement: the side edge of the paint between the low post and the free-throw line.\par
\textbf{11.000\textendash{}12.833~s}\enspace Ryan Dunn attempted a three-point shot. Location / Movement: the corner three-point area.\par
\textbf{12.833\textendash{}14.667~s}\enspace Devin Booker secured a rebound. Location / Movement: the landing area of a long rebound in the high area outside the free-throw line.\par
\textbf{15.333\textendash{}18.700~s}\enspace Devin Booker is dribbling the ball.\par
\end{rubricpanel}

\newpage\vspace*{-17pt}\begin{rubricpanel}{Offensive review (R1)}
\textbf{Coverage:} Full possession.\par
\smallskip
\textbf{0.000\textendash{}0.833~s}\enspace Ryan Dunn inbounded the ball. Location / Movement: out of bounds behind the baseline beneath the backcourt basket. Recorded result: success.\par
\textbf{1.000\textendash{}7.000~s}\enspace Tyus Jones dribbled the ball. Location / Movement: advancing to the top of the arc in the frontcourt. Recorded result: success.\par
\textbf{7.000\textendash{}8.000~s}\enspace Tyus Jones passed the ball. Location / Movement: beyond the three-point line at the top of the arc. Recorded result: success.\par
\textbf{8.000\textendash{}9.667~s}\enspace Devin Booker dribbled the ball. Location / Movement: driving inward after receiving the ball on the perimeter. Recorded result: success.\par
\textbf{10.167\textendash{}11.000~s}\enspace Devin Booker passed the ball. Location / Movement: the side edge of the paint between the low post and the free-throw line. Recorded result: success.\par
\textbf{11.000\textendash{}12.833~s}\enspace Ryan Dunn attempted a three-point shot. Location / Movement: the corner three-point area. Recorded result: missed.\par
\textbf{12.833\textendash{}14.667~s}\enspace Devin Booker secured a rebound. Location / Movement: the landing area of a long rebound in the high area outside the free-throw line. Recorded result: offensive.\par
\textbf{15.333\textendash{}20.167~s}\enspace Devin Booker dribbled the ball. Location / Movement: returning to the perimeter to organize the offense, dribbling in the high area. Recorded result: success.\par
\textbf{20.333\textendash{}21.333~s}\enspace Devin Booker passed the ball. Location / Movement: inside the arc and just outside the free-throw line. Recorded result: success.\par
\textbf{21.333\textendash{}22.167~s}\enspace Oso Ighodaro attempted a dunk. Location / Movement: under the basket. Recorded result: success.\par
\end{rubricpanel}

\subsection{Qualitative Analysis}

\noindent\textbf{Error Analysis.}
We sampled 810 prediction explanations and 360 retrospective answers from six model configurations. For each response, we obtained a mean score $s$ on a scale from 1 to 5 by averaging its scores across the corresponding evaluation dimensions. We used three dimensions for prediction explanations and four for retrospective answers. We defined low scores as $s\le 3$, medium scores as $3<s\le 4$, and high scores as $4<s<5$. The target sampling proportions for these three score bands were 40\%, 40\%, and 20\%, respectively. We manually assigned each response a label corresponding to its primary error.

\begin{enumerate}
\item \textbf{Visual Perception Error.} The model observes the relevant footage but misidentifies a player or misinterprets an action, a position, or the order of events.
\item \textbf{Visual Hallucination.} The model describes actions, player involvement, or offensive or defensive plays absent from the input footage as observed facts.
\item \textbf{Task and Domain Understanding Error.} The model misunderstands the question or misapplies basketball rules, concepts, or terminology.
\item \textbf{Reasoning Error.} The model identifies the relevant facts but draws an incorrect inference about a causal
\end{enumerate}

\onecolumn
\noindent
\begin{minipage}[t]{\dimexpr(\textwidth-\columnsep)/2+6pt\relax}
    \vspace{0pt}
    \begin{enumerate}
        \setcounter{enumi}{4}
        \item[] relationship, a tactical mechanism, or what happens next.
        \item \textbf{Missing Key Information.} The model omits evidence needed to support the prediction or explain the play.
    \end{enumerate}
\end{minipage}\hfill
\begin{minipage}[t]{\dimexpr(\textwidth-\columnsep)/2\relax}
    \vspace{0pt}
    \noindent\textbf{Visualization.}
    Figure~\ref{fig:qualitative_cases} presents six cases. The first case shows a successful action prediction. Each of the five remaining cases illustrates a different error category.
\end{minipage}

\par\medskip
\noindent
\begin{minipage}{\textwidth}
    \centering
    \captionsetup{type=figure}
    \includegraphics[width=\linewidth,height=0.865\textheight,keepaspectratio]{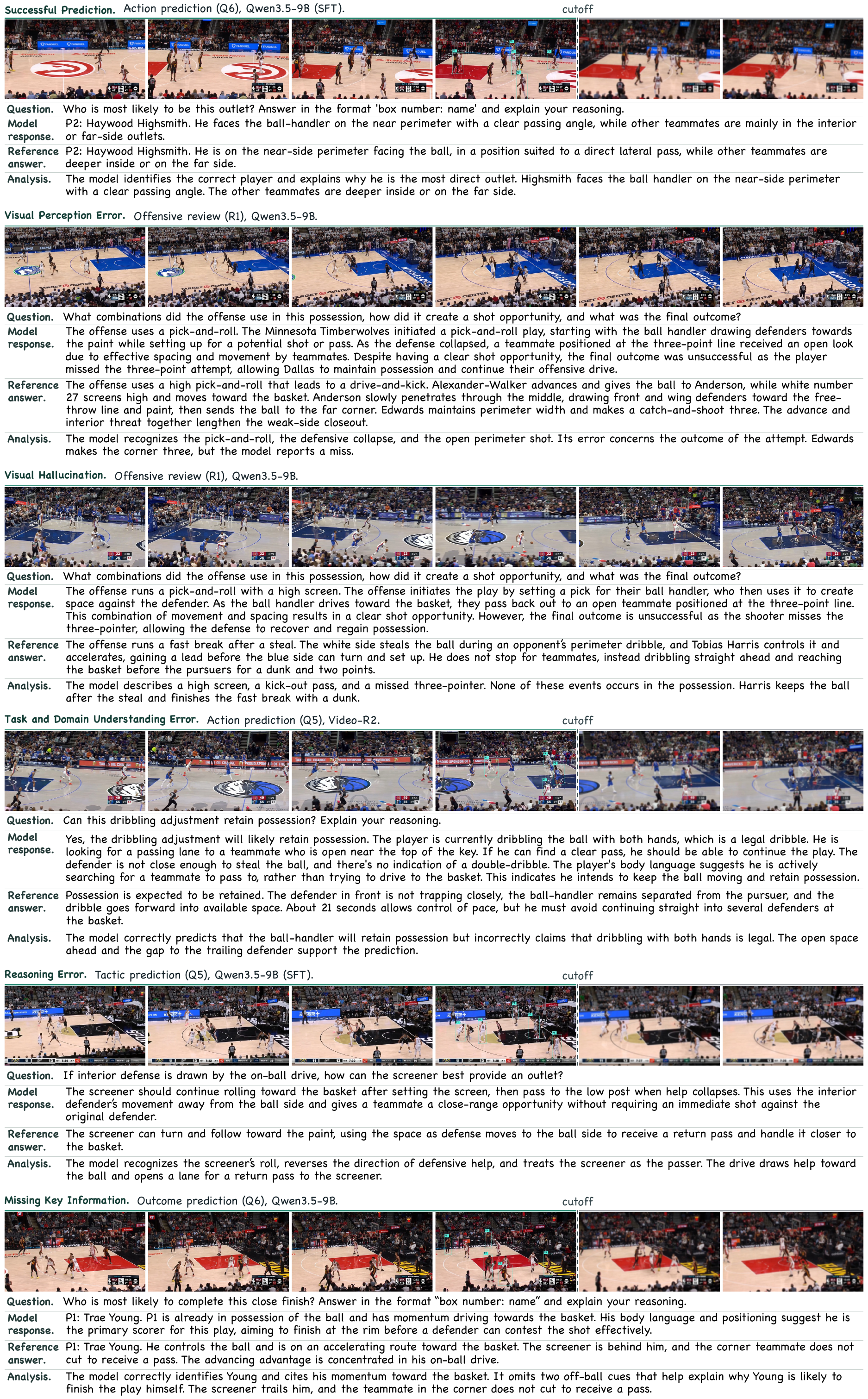}
    \caption{Qualitative examples from BasketTactics. The first case shows
    a successful prediction. The remaining five cases illustrate one
    example from each error category.}
    \label{fig:qualitative_cases}
\end{minipage}

\end{document}